\documentclass[runningheads]{llncs}
\usepackage{graphicx}
\usepackage{url}  %Required
\usepackage{graphicx}  %Required
\usepackage[nolist]{acronym}
\usepackage{subfigure}
\usepackage{amsmath} % assumes amsmath package installed
\usepackage{amssymb}  % assumes amsmath package installed

\usepackage{dblfloatfix}

\begin{document}

\title{Reasoning on Grasp-Action Affordances}

\author{Paola Ard\'on\orcidID{0000-0002-3026-0706} \and
\`Eric Pairet\orcidID{0000-0002-3363-0426} \and
Ron Petrick\orcidID{0000-0002-3386-9568} \and
Subramanian Ramamoorthy\orcidID{0000-0002-6300-5103} \and
Katrin Lohan\orcidID{0000-0001-9843-316X}}
\authorrunning{P. Ard\'on et al.}
\institute{Edinburgh Centre for Robotics. Edinburgh, UK. \\
\email{paola.ardon@ed.ac.uk}}

\maketitle

\begin{acronym}[ransac]
  \acro{LbD}{learning by demonstration}
  \acro{RL}{reinforcement learning}
  \acro{SVM}{support vector machine}
  \acro{DOF}{degrees-of-freedom}
  \acro{CAD}{computer-aided design}
  \acro{ROI}{regions of interest}
  \acro{MCMC}{Markov Chain Monte Carlo}
  \acro{ECV}{early cognitive vision}
  \acro{IADL}{instrumental activities of daily living}
  \acro{CDR}{cognitive developmental robotics}
  \acro{2-D}{two-dimensional}
  \acro{3-D}{three-dimensional}
  \acro{RANSAC}{Random sample consensus}
  \acro{RGB-D}{red-green-blue depth}
  \acro{IFR}{International Federation of Robotics}
  \acro{CNN}{Convolutional Neural Networks}
  \acro{KB}{Knowledge Base}
  \acro{MSE}{mean square error}
\end{acronym}

%\tableofcontents
\vspace{-0.5cm}
\begin{abstract}
Artificial intelligence is essential to succeed in challenging activities that involve dynamic environments, such as object manipulation tasks in indoor scenes. Most of the state-of-the-art literature explores robotic grasping methods by focusing exclusively on attributes of the target object. When it comes to human perceptual learning approaches, these physical qualities are not only inferred from the object, but also from the characteristics of the surroundings. This work proposes a method that includes environmental context to reason on an object affordance to then deduce its grasping regions. This affordance is reasoned using a ranked association of visual semantic attributes harvested in a knowledge base graph representation. The framework is assessed using standard learning evaluation metrics and the zero-shot affordance prediction scenario. The resulting grasping areas are compared with unseen labelled data to asses their accuracy matching percentage. The outcome of this evaluation suggest the autonomy capabilities of the proposed method for object interaction applications in indoor environments.
\end{abstract}
\vspace{-0.5cm}
%This work proposes a method that includes environmental context to reason on an object affordance which informs the selection of the grasping regions

%%%%%%%%%%%%%%%%%%%%%%%%%%%%%%%%%%%%%%%%%%%%%%%%%%%%%%%%%%%%%%%%%%%%%%%%%%%%%%%%
% Sections
\section{Introduction}\label{sec:intro}

%Affordances in general
One of the most significant challenges in artificial intelligence is to achieve a system that simulates human-like behaviour.
Let us consider a robot in a simple task such as finding, collecting and delivering an object in home environments. Given the complexity of home settings, it is hard to provide a robot with every possible representation of the objects contained in a house. It is even harder to feed the robot with all the possible uses of those objects. Instead of learning all possible scenarios, suppose that a reasoning technique allows the system to deduce an object affordance. As a result, offering the opportunity to achieve autonomous capabilities. The term affordance refers to everything that defines the interaction with an object, from the way to grasp it to its inherited ability to perform different tasks~\cite{Gibson77-affordances}. Thus, affordance defines all possible actions depending on the target objects' physical capabilities. For instance, a glass cup looks as if it can be handed over, contain liquids, or pour liquids from it. The characteristics that define the glass cup as a container or graspable object constitute its affordance.
%%%%%%%%%%%%%%%%%%%%%%%%%%%%%%%%%%%%%%%%%%%%%%%%%
\begin{figure}[t]
  \centering
  \includegraphics[width= 8.5cm]{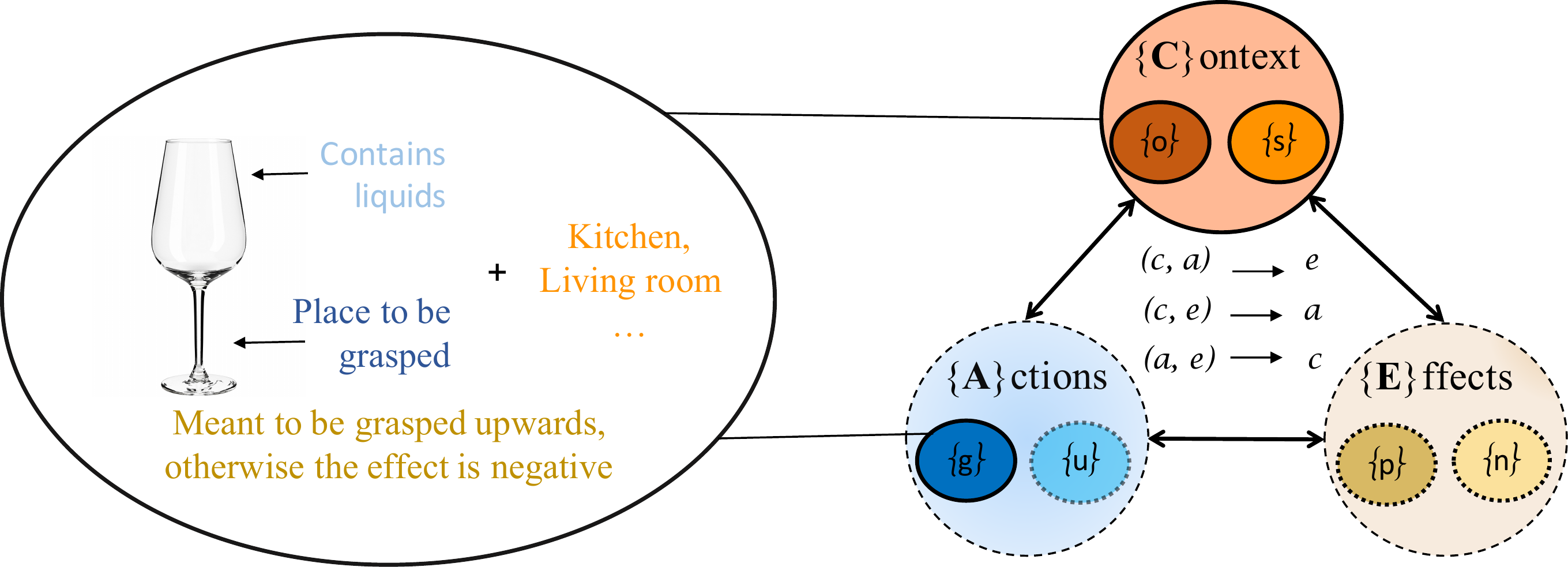}
  \caption{Affordance map model to create a correlation between the objects properties and their environment to improve on grasp-action affordance.\label{fig:affordances_components}}
  \vspace{-0.5cm}
\end{figure}
%%%%%%%%%%%%%%%%%%%%%%%%%%%%%%%%%%%%%%%%%%%%%%%%%
%In robotics
According to different theories of human perception, the psychology of perceptual learning compounds the different qualities in the environment rather than acquiring associated responses to every object~\cite{gibson2014ecological,de2008perceived}. Thus, humans are efficient at deducing affordance for objects with different appearances and similar abilities,  e.g. glasses: wine, tumbler, martini, and discern among those with similar features but different purposes, e.g. bowling pin vs water bottle.
Nonetheless, in robotics, the most common approach to learn affordances is from labels \cite{Montesano2008LearningOA,Lenz2015,bonaiuto2015learning}. This technique limits the number of learned objects, grasping areas and affordance groups. Moreover, the robot is unable to interact with novel objects. Further, by learning the limited set of responses, it is not possible to deduce the key features that define the objects affordance. 

Using the same analogy as the theories of human perception this paper hypothesises that using the semantic features of the object and its surroundings not only improves the affordance grasping action towards the object but it also allows a reasoning process that, in the long term, offers autonomy capabilities, a solution not yet seen in the current literature. 
%This work
This work summarises an architecture that addresses the previously described challenges. The focus is on affordance reasoning for calculating grasping areas, using a combination of the object and its environment features. Figure~\ref{fig:affordances_components} shows the foundations of this proposal, which is an extended version of the affordance map presented in \cite{Montesano2008LearningOA}. The proposed methodology works with the concept that an affordance relates attributes of an object and the environment to an interactive activity by an agent who has some ability, which relates back to the object causing some affordance. In other words, the attributes of the object and the environment reside in the context of the affordance, the abilities of the agent and the object in the affordance actions and the outcome of this interactive activity in the effects. This work focuses on the integration of the semantic features of the previously mentioned environment in order to obtain a good grasp affordance action, from now on referred to as grasp-action, of the object.
The presented framework can reason on the object grasping areas that are strongly related to the affordance group. The reasoning process is based on a \ac{KB} graph representation. This \ac{KB} is built using semantic attributes of the object and the environment. For every object explored by the framework, the \ac{KB} uses weights to relate a subset of attributes. This association then leads to an affordance category which is highly correlated with a grasp-action area. The designed framework is assessed not only using standard learning evaluation metrics, but it is also tested on the zero-shot affordance prediction scenario. Moreover, the resulting grasping areas are compared with unseen labelled data to asses their accuracy matching percentage. The results demonstrate the suitability of the method for grasp-action affordance applications, offering a generalised object interaction alternative with autonomy capabilities.
\section{Related Work}\label{sec:related_work}

Many methods extract viable grasping points on objects, independently on their affordance \cite{Lenz2015,ardon2018Reaching}. Others focus explicitly on the task of grasp-action affordance from visual features and model parameters that are learned through reinforcement learning using biologically inspired methods \cite{stoytcvhev2005toward,bonaiuto2015learning}. \cite{stoytcvhev2005toward}, interestingly embraces psychology theories for human development such as the ones presented in \cite{gibson2014ecological} to learn from exploratory behaviours the invariants to obtain the best grasps. 
Contrary, \cite{ardon2018HRI,moldovan2012learning}
focus on the ability-action affordance of the objects. In their work, they use statistical relational learning to learn the ability affordance of different objects, which shows to cope with uncertainty.
Other works go beyond the visual representation of the object and combine visual as well as textual descriptors to build a \ac{KB} \cite{zhu2014reasoning,Sridharan}. This \ac{KB} is composed of actions learned through reinforcement learning techniques with the purpose of interacting with the object. \cite{geib2006object,kruger2011object} work on the actions and objects relations in a single interface representation to capture the needs of planning and robot control. Another extension is \cite{detry2009learning}, they use these action complexes to extract the best grasping points of the objects. In literature, it is extensive the use of learning techniques such as deep \ac{CNN} to build an affordance model based on the visual objects features, resulting in a plausible generalised method given the robustness of their data \cite{Nguyen17,AffordanceNet18}. Unlike these works, this paper presents a methodology that combines attributes of the object and the environment to provide a denser context for object affordance interaction. Thus, allowing it to generalise the grasp-action affordance on similar objects. 

\section{Proposed Solution \label{sec:method}}

\begin{figure}[t!]
  \centering
  \includegraphics[width=8.7cm]{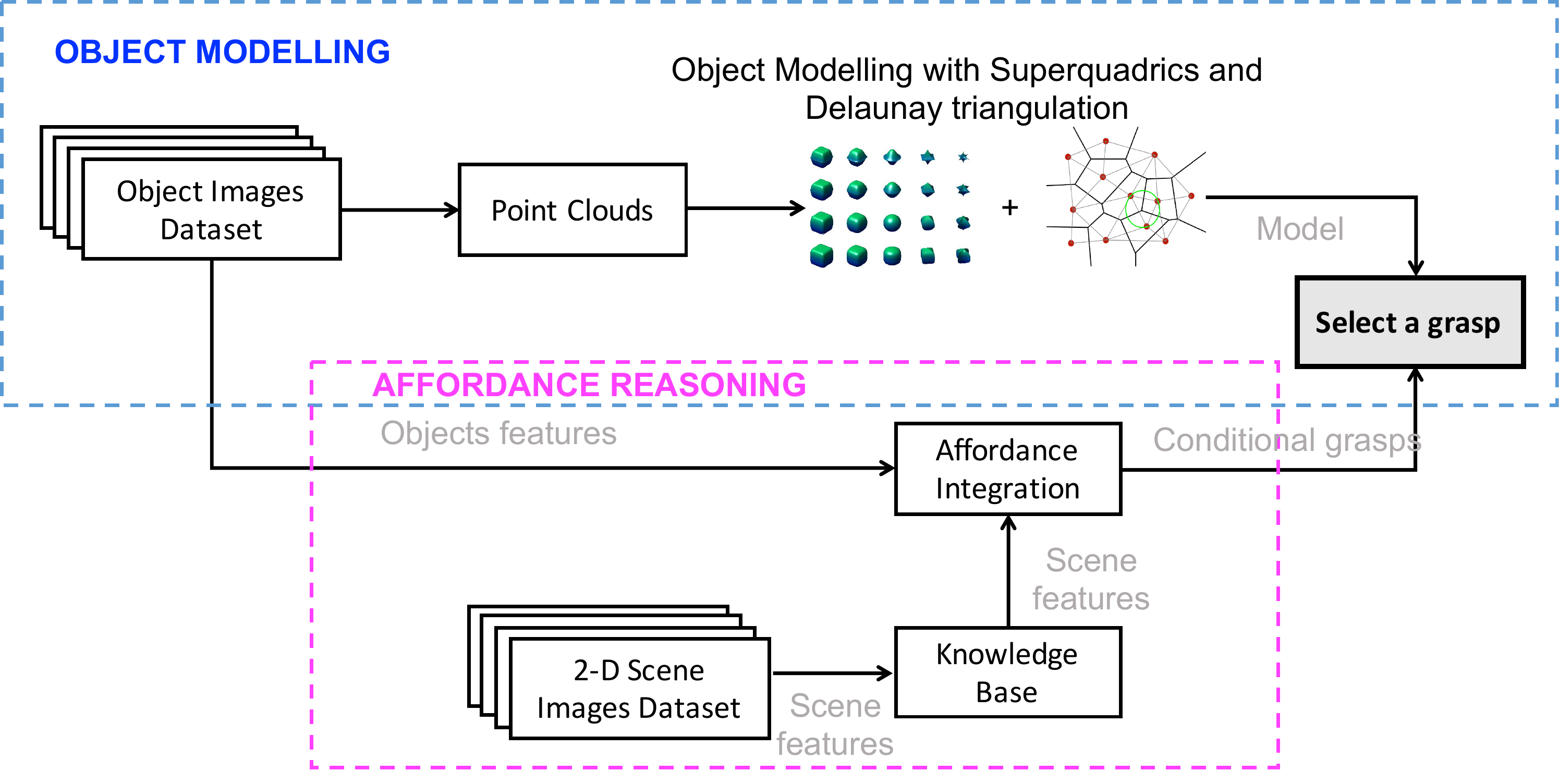}
  \caption[Proposed framework for grasping affordance inference.]{Proposed framework for grasping affordance reasoning.}
  \label{fig:complete}
  \vspace{-0.5cm}
  \end{figure}

In this paper, a grasp-action area of the object is the result of the relation between the object and its surrounding environment. Figure~\ref{fig:affordances_components} shows a summary of the proposed affordance model. Let us consider a glass cup in an affordance map relationship. Additional to its inherited affordance action qualities, i.e. contain liquids and being graspable, there are other elements that define its opportunity of interaction. For example the way in which it is being manipulated as well as the features that describe the glass cup itself and its surrounding environment. All these elements together define the affordance of the glass cup. This does not mean they are dependant of each other but rather codefining and coherent together. Bearing this example in mind, in Figure~\ref{fig:affordances_components}, the context \mbox{\textbf{C}~$ = \{c_1, c_2, ..., c_n\}$} is the set of semantic attributes of the glass cup and its environment (such as kitchen and living room), \mbox{(\{o\}bject $\cup$ \{s\}urrounding) $\subseteq$ \{\textbf{C}\}ontext}. The set of available actions, \mbox{\textbf{A}~$ = \{a_1, a_2, ..., a_n\}$}, is understood as a twofold: (i)~the way in which the glass cup can be approached, its suitable grasp-action areas, and~(ii)~the usages that the glass cup can achieve, its ability-action such as containing liquids, \mbox{(\{g\}rasps $\cup$ \{u\}sage) $\subseteq$ \{\textbf{A}\}ctions}. The set of effects of performing those actions, \mbox{\textbf{E}~$ = \{e_1, e_2, ..., e_n\}$}, is kept as a simple discretisation between positive or negative effects, such as holding the glass cup correctly in order not to spill the liquid, \mbox{(\{p\}ositive $\cup$ \{n\}egative) $\subseteq$ \{\textbf{E}\}ffects}.
The key attributes of the affordance reasoning to get those grasp-action areas are enclosed in the form of a \ac{KB}.
These methods are commonly used in artificial intelligence because of their advantages for harvesting data and accessing a more extensive array of queries regarding the essential features of a process, rather than just the result. \acp{KB} achieve this task by connecting a collection of attributes through a general set of rules. In this work, the attributes are the features that describe the object and the environment and are connected through a hierarchical set of decisions that result in the object affordance. This section first summarises the object modelling stage, to then reason on the object affordance that is highly correlated with the resulting grasp-action areas as schematised in Figure~\ref{fig:complete}.
%--------------------------------------   
%---------------------------------------

%--------------------------------------
%--------------------------------------
%--------------------------------------
% \subsection{Pre-prosessing\label{sc:modelling}}
% This stage consists of the segmentation of the target objects from \ac{3-D} point clouds to obtain a model that helps on defining its grasping areas.
% As done in \cite{ardon2018HRI}, this framework combines superquadric object reconstruction techniques \cite{jaklic2013segmentation} with Delaunay triangulation \cite{lee1980two} to obtain a \ac{3-D} model of the object. The points obtained from the Delaunay triangulation go through a threshold and clustering process. These different clustered areas are then considered as the preliminary grasping regions. Even though this solution is not an accurate representation of the object, it allows the framework to model and extract grasping points without any \textit{a-priori} acknowledge about the target.

%--------------------------------------
%--------------------------------------
%--------------------------------------
% \subsection{Knowledge Base Components\label{sc:components}}
%   \begin{figure}[b!]
%   \centering
%   \includegraphics[width=8.5cm]{Figures/kb} \caption[Example of the different entities in the \ac{KB}.]{Example of the different entities in the \ac{KB}.\label{fig:kb_example} }
% \end{figure}
%------------------------------
 A \ac{KB} is visualised as a graph representation, as illustrated in Figure~\ref{fig:affordances_to_clean_example} where the entities (nodes) are connected by general rules (edges). In this setup, the entities are the target object, the attributes of the object and its surrounding, and the resulting affordance groups. The general rules are the attribute to attribute relation that results from a classification process.  
 The relation between attributes are weighted accordingly, where the higher the weight, the higher the correlation between the two entities.
In order to describe objects by their attributes the best practice is to divide their features into base, semantic and discriminative \cite{farhadi2009describing}.
 In this work, the base features, such as edges and colours, are extracted using \ac{CNN}. The semantic features are visual characteristics of the object as defined in Table~\ref{tb:classifiers}. From now on, these features will be referred to as visual semantic features. They are the result of a deep \ac{CNN} and are divided as (i)~shape attributes, these are the set of visual attributes that describe the objects geometrical appearance; (ii)~texture attributes, are categories based on visual characteristics of the objects materials; and (iii)~environment attributes, which are the scenarios in which the objects are more likely to be found in. This attribute is added with the purpose of facilitating the object affordance reasoning. The implemented \ac{KB} considers different scenarios in which the object can be located; thus the object is not restricted to a particular environment. For example, a glass containing liquids is more likely to be found in a kitchen and a living room. 
 Finally, the discriminative features are those that offer a comprehensive understanding of the semantic features. They are the result of a predictive decision tree model that uses deep \ac{CNN} as nodes.
 %-----------------------------
\begin{table}[t!]
    \centering
    \begin{tabular}{|c|c|c|}
    \hline
    \textbf{Attribute}              & \textbf{Entities per Attribute} \\
    \hline
    Shape  &  box, cylinder, irregular, long, round  \\ \hline
    Texture            & \begin{tabular}{c}aluminium, cardboard, coarse, \\fabric, glass, plastic, rubber, smooth\end{tabular} \\ \hline
    Categorical & \begin{tabular}{c} container, food, personal,\\ miscellaneous, utensils\end{tabular} \\ \hline
    Environment & \begin{tabular}{c} bathroom, bedroom, play-room,\\ closet, kitchen, living room, office\end{tabular} \\ \hline
    \end{tabular}
    \caption{Used attributes and entities of the \ac{KB} graph.\label{tb:classifiers}}
    \vspace{-0.5cm}
\end{table}
%-----------------------------
The \ac{KB} is composed of four different Deep Neural Networks that, through the pre-trained \ac{CNN}, resnet50~\cite{he2016deep}, extract features from the perceived images. These four different deep learned \ac{CNN} correspond to the four different visual semantic attributes, as described in Table~\ref{tb:classifiers}, which result in the deduced set of entities in a graph that defines a grasp-action affordance.

\vspace{-0.3cm}
%--------------------------------------
%--------------------------------------
%--------------------------------------
 \subsection{Knowledge Base Predictive Model\label{sc:affordances}}
 \vspace{-0.3cm}
 %----------------------------
  \begin{figure}[b!]
  \vspace{-0.5cm}
  \centering
\includegraphics[width=7cm]{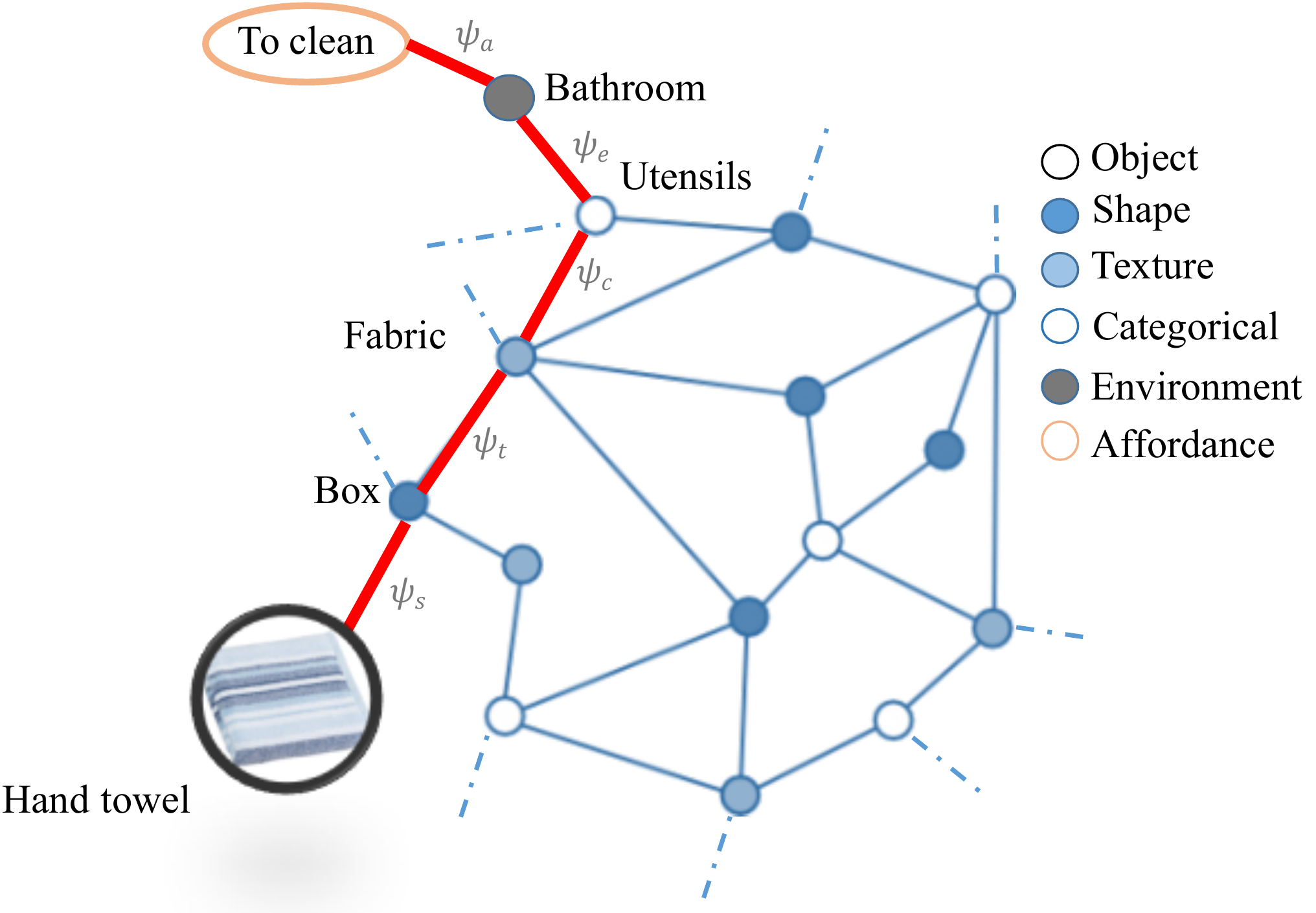}
 
  \caption[Example of a cleaning object and the extracted attributes used to build the \ac{KB} graph.]{Example of a cleaning object and the extracted attributes used to build the \ac{KB} graph. The higher weights $\boldsymbol{\Psi}$ (red) create the reasoning to an affordance group.\label{fig:affordances_to_clean_example} }
\vspace{-0.5cm}
\end{figure}
%------------------------------

In this paper, the \ac{KB} constitutes a data library that builds a predictive model connected through a hierarchical set of decisions, such as the edges on Figure~\ref{fig:affordances_to_clean_example}, from now on referred as weights. These decisions are the result of a classification task of the object semantic features, represented as the nodes in Figure~\ref{fig:affordances_to_clean_example}. From each of the attributes, ${\forall a \in \boldsymbol{A} : \boldsymbol{A} \in [1, ..., K]}$, where $K$ is the total number of visual semantic features as described in Table~\ref{tb:classifiers}, a set of weights represented as a vector ${\boldsymbol{\Psi_{a_k}} = [\psi_1, \psi_2, ..., \psi_n]}$ is extracted, where $n$ is the total number of entities in that attribute. These $\boldsymbol{\Psi_{a_k}}$ are hierarchically connected with the next attribute ${a_{k+1}}$. Then ${\boldsymbol{\Psi_{a_k}}}$ offers a way to rank on the next best entity candidate. The higher the $\psi_n$, the higher the probability that the connected two entities among attributes result in a better affordance reasoning. These weights are proportional to the posterior probability distribution obtained from the classification task. Such that the posterior probability distribution is defined as the Bayes rule:
 \begin{equation}
 \widehat{P}(a|x) = \frac{P(x|a) P(a)}{P(x)},
 \end{equation}
where $x$ is an image belonging an attribute $a$, $P(a)$ is the posterior distribution and $P(x)$ is a normalisation constant that consists of the sum over $a$ of the multivariate normal density. Figure~\ref{fig:affordances_to_clean_example} depicts an example of an object which grasping affordance can be to clean or to hand over. In this example, the weights deduce the best path (shown in red) to the \textit{to clean} grasping affordance. The collected information from each of the deep \ac{CNN} is then used to learn a decision tree as a predictive model:
$
(\boldsymbol{y}, Z) = (y_{1},y_{2},y_{3},...,y_{n},Z),
$
where $Z$ is the affordance group that the system is trying to reason, and the vector $\boldsymbol{y}$ is the set of features $\{y_{1},y_{2},y_{3},...,y_{n}\}$ used for the reasoning task.
Thus, the model learns the ranking that reasons on the affordance grasping task ${R(x) = \boldsymbol{\Psi^{\intercal}_{A}} \boldsymbol{y}(x)}$ where $\boldsymbol{\Psi_{A}}$ is the transpose of the model parameters from all the attributes and $\boldsymbol{y}(x)$ is the set of visual features of a given image $x$.
\vspace{-0.3cm}
%--------------------------------------
%--------------------------------------
%--------------------------------------
\subsection{Calculating the Grasping Points}
\vspace{-0.3cm}
Once the affordance is deduced, the system selects from the set of grasping points obtained in the object reconstruction stage and limits the grasps depending on the affordance reasoning obtained from the \ac{KB}. In order to impose such constraints, the space of the previously obtained grasping points is discretised in the third dimension, $z$, so that the following decision on the grasping area can be made: (i)~The grasping region should lie on those points located in the central subspaces of the discretised space for objects that are meant to contain edibles. (ii)~For the rest of objects, it is considered as the grasping region those subspaces where the density of grasping points is higher than a threshold, given that the affordance action-effect is not critical.
% \begin{itemize}
%   \item The grasping region should lie on those points located in the central subspaces of the discretised space for objects that are meant to contain edibles.
%   \item For the rest of objects, it is considered as the grasping region those subspaces where the density of grasping points is higher than a threshold, given that the affordance action-effect is not critical (i.e., hand over, to clean, among others.).
% \end{itemize}

\section{Evaluation}\label{sec:results}

This work's goal is to achieve a system able to reason on the object grasp-action affordance, thus offering autonomy capabilities. As a result, it is of interest to evaluate the \ac{KB} on (i)~its attribute accuracy classification, and (ii)~its reasoning efficiency with similar objects.

% The proposed method's goal is to achieve a system with human-like object interaction capabilities, thus it is of interest to evaluate the \ac{KB} on (i)~the accuracy of the classification per attribute, (ii)~its inference efficiency with similar objects, (iii)~its discerning capabilities regarding the affordance of semantically similar objects, and (iv)~its performance compared with similar methodologies.
\vspace{-0.3cm}
%--------------------------
%--------------------------
%--------------------------
\subsection{System Setup\label{sc:collecting}}
\vspace{-0.3cm}
% %----------------------------------------
% \begin{figure}[t!]
%     \centering
%   \includegraphics[width=7cm]{Figures/Affordances_classification_and_grasps}
%     \caption[Sample of objects used for the framework from the Washington and MIT datasets and the different affordances groups.]{Sample of images used for the \ac{KB}. The columns are objects from the Washington-RGB dataset and the rows scenes from the MIT dataset. The bottom of the table depicts the resulting affordance  categories.\label{fig:affordances_classification}}
%   \end{figure}
% %------------------------------------------
The setting up of the system consists on collecting the required data for the training and the assessment of the method. This collection is built using two different datasets that are manually organised into entities of the attributes described in Table~\ref{tb:classifiers}. After passing through the predictive model in the \ac{KB}, every object in the library is expected to fall into: \textit{to eat, to contain, to hand over, to brush, to squeeze, to clean} or \textit{to wear}. The first set of images is from the Washington-RGB dataset, which contains $300$ objects providing the point clouds and the \ac{2-D} images for each one of the instances~\cite{lai2012detection}. 
The second dataset is the MIT indoor scene recognition that contains $15{,}620$ different \ac{2-D} images of $67$ different indoor scenes from which this work uses seven of those classes \cite{quattoni2009recognizing}. By unifying these two datasets, the objects are correlated to the environment in which they are more likely to be located.
%-------------------------
%-------------------------
%-------------------------
   \begin{figure*}[b!]
   \vspace{-0.5cm}
      \centering
        \subfigure[ ]{\label{fig:confusion_before} \includegraphics[width=5.5cm]{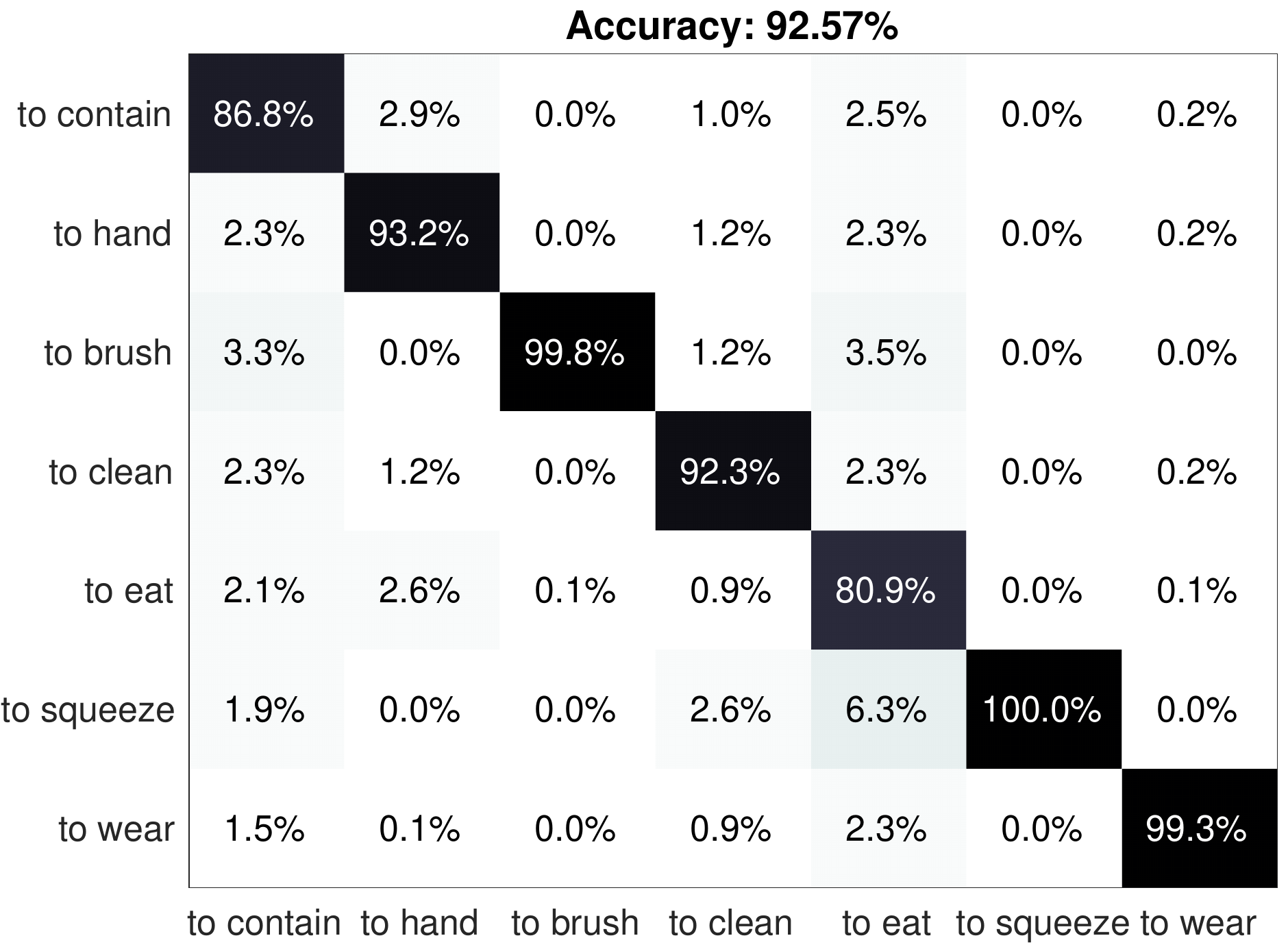}}
        \subfigure[ ]{\label{fig:confusion_after} \includegraphics[width=5.5cm]{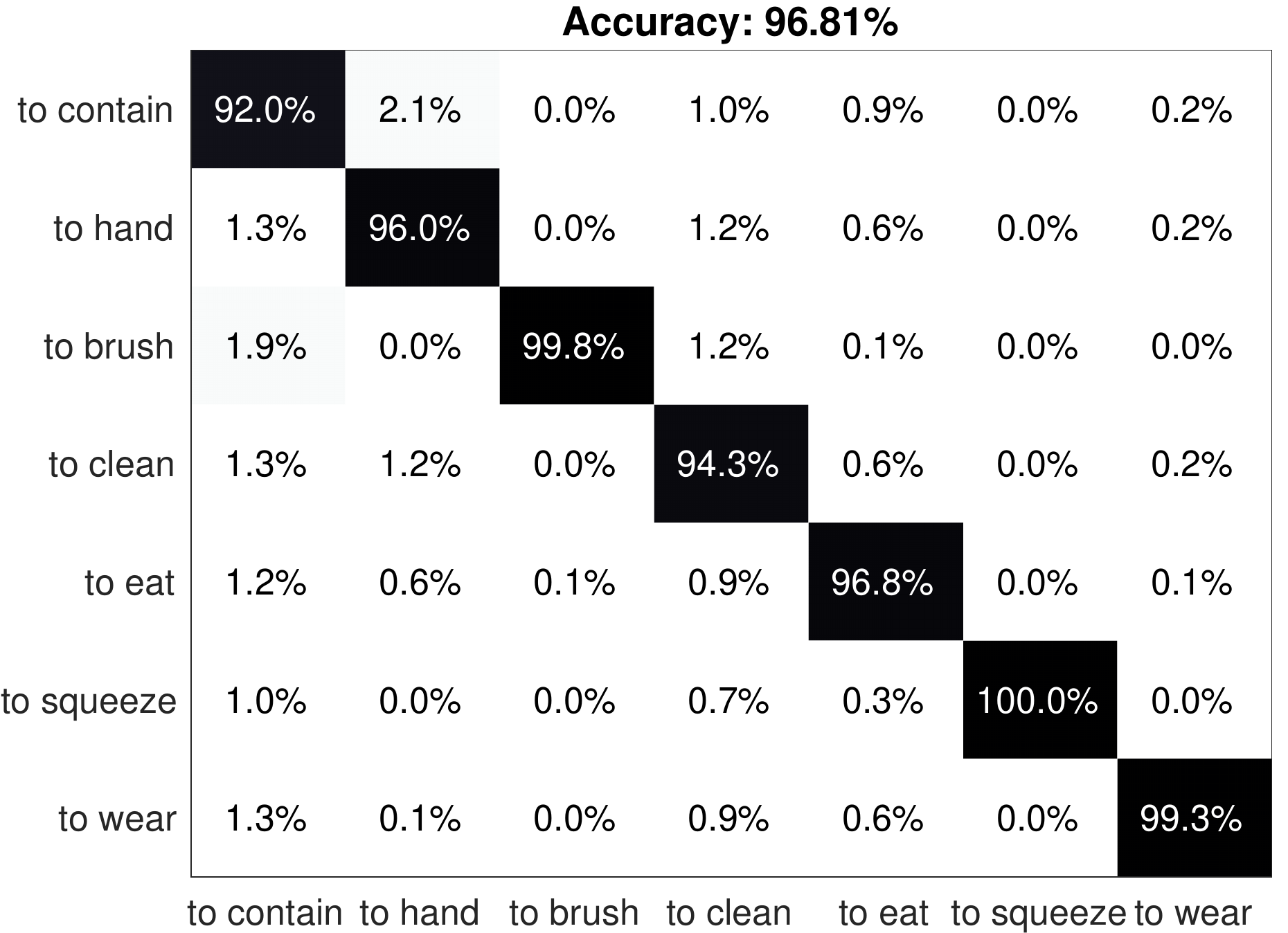}}
        \caption[Affordance category classification performance.]{Affordance category classification performance:~(a)~before adding environment features, showing an average diagonal accuracy of 92.57\%;~(b)~after including the environment, showing a diagonal average accuracy of 96.81\%.}
        \label{fig:confusion_matrices}
        \vspace{-0.5cm}
    \end{figure*}
%-------------------------
%-------------------------
%-------------------------
Both datasets are split into $70\%$ for training and the remaining $30\%$ for testing. These subsets are used to train and test a battery of classifiers that help to define good object affordances features. In order to represent the obtained grasping area of the objects, an ellipse with the iCub humanoid robot end-effector dimensions is simulated. The orientation of such ellipse is out of the scope of this work and the focus remains on the position of the grasping area. 
\vspace{-0.3cm}
%--------------------------
%--------------------------
%--------------------------
\subsection{Reasoning on the Affordance}
\begin{figure}[t!]
    \centering
   \includegraphics[width=7cm]{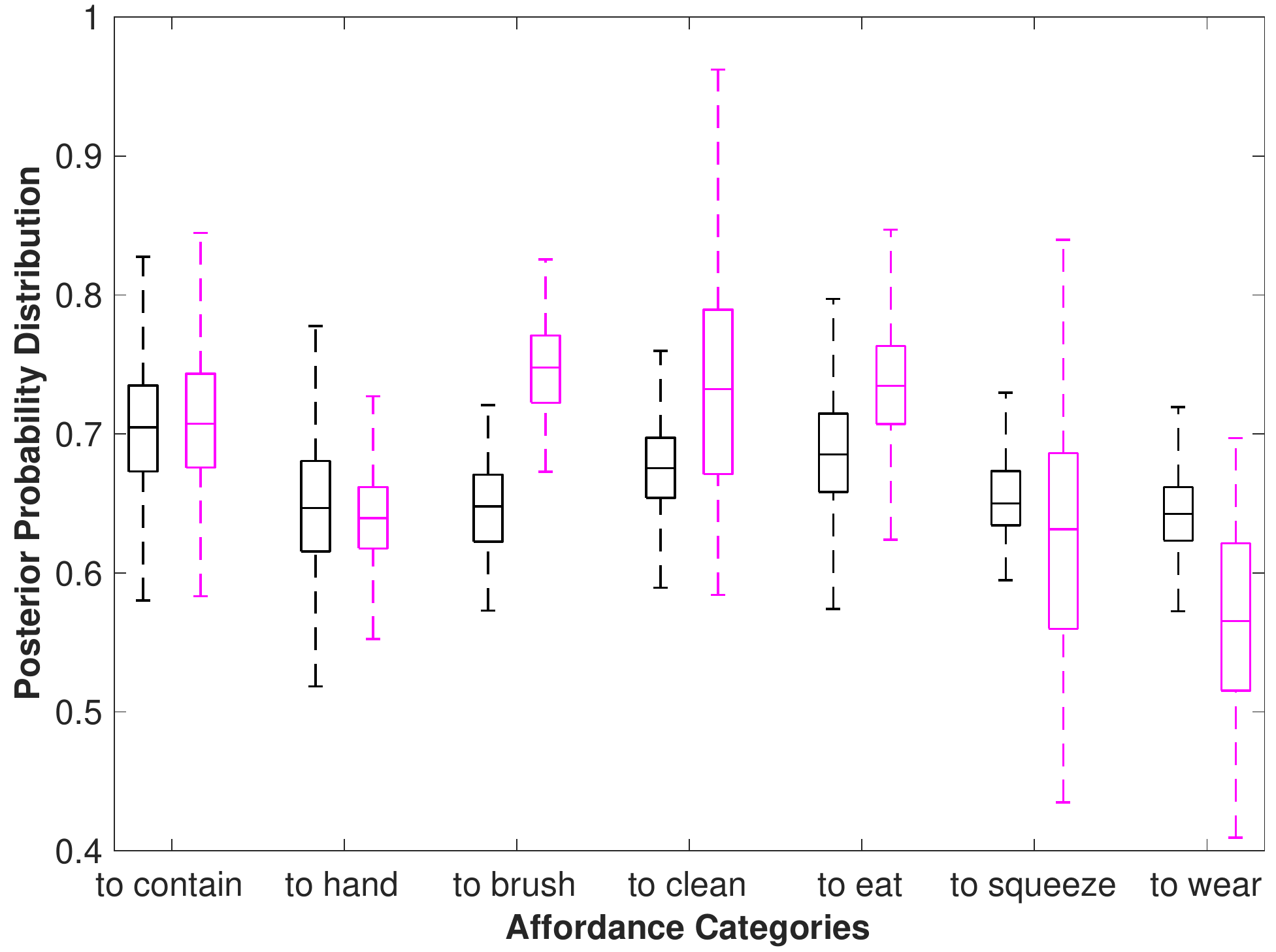}
    \caption[Distributional posterior probabilities per class of the knowledge base.]{Distributional posterior probabilities per class of the \ac{KB} before (shown in black) and after (shown in magenta) the environment features.\label{fig:box}}
    \vspace{-0.5cm}
  \end{figure}

\vspace{-0.3cm}
A summary of the accuracies per deep \ac{CNN} in the \ac{KB} is presented in Table~\ref{tb:classifiers_performance}.
As a reminder to the reader, the aim of the proposed methodology is not to improve the performance of the individual classifiers. Nonetheless, the illustrated accuracies match the state-of-the-art results shown in~\cite{he2016deep,lai2012detection}.
%-------------------------
\begin{table}[t!]
    \centering
    \begin{tabular}{|c|c|c|}
    \hline
    \textbf{Classifier}              & \textbf{Accuracy}  \\
    \hline
    Shape  &  95.71\%   \\ \hline
    Texture            & 98.83\% \\ \hline
    Categorical & 99.91\% \\ \hline
    Environment & 76.50\% \\ \hline
    \end{tabular}
    \caption{Each of the attributes classification accuracies.\label{tb:classifiers_performance}}
    \vspace{-0.5cm}
\end{table}
%-------------------------
To evaluate the overall performance of the \ac{KB}, the accuracies before and after adding the environment features were collected.
Figure~\ref{fig:confusion_matrices} shows the data for both cases. A lower accuracy is obtained in the case where the environment features are not included, as illustrated in Figure~\ref{fig:confusion_before} and Figure~\ref{fig:confusion_after}. Furthermore, Figure~\ref{fig:confusion_before} not including the environment shows a slightly higher spread among different affordance classes. This misclassification is the case for affordances which objects have a general semantic categorical attribute such as ``miscellaneous'' or ``container''. Thus, a percentage of objects are misclassified among the \textit{to contain, to brush, to eat,} and \textit{to squeeze} categories. Regarding grasping, this miscue represents a significant adverse effect, especially for objects which real affordance is \textit{to contain}, and its misclassification results in the system lifting up the object from any point, risking dropping its content. 
This risk is reduced by $4.24\%$ when adding the environment features, as portrayed in Figure~\ref{fig:confusion_after}, especially in categories such as \textit{to contain, to hand over} and \textit{to eat}. 
The posterior probability distribution of the affordances categories is also evaluated. Figure~\ref{fig:box} shows that 
while there is a decrement in the distribution for some categories such as \textit{to hand}, there is an increment for others such as \textit{to clean}. This change in the distribution is accredited to the variation in environments where these objects are found. 
%-------------------------

\vspace{-0.3cm}
%----------------------------
%----------------------------
%----------------------------
\subsection{Zero-shot Affordance\label{sc:zero}}
\vspace{-0.3cm}

%----------------------
\begin{figure*}
        \centering
        \subfigure[to hand over]{
            \begin{minipage}{0.18\linewidth}
                \centering
                \includegraphics[width=2cm]{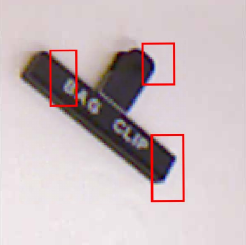}
                
                \includegraphics[width=1.8cm]{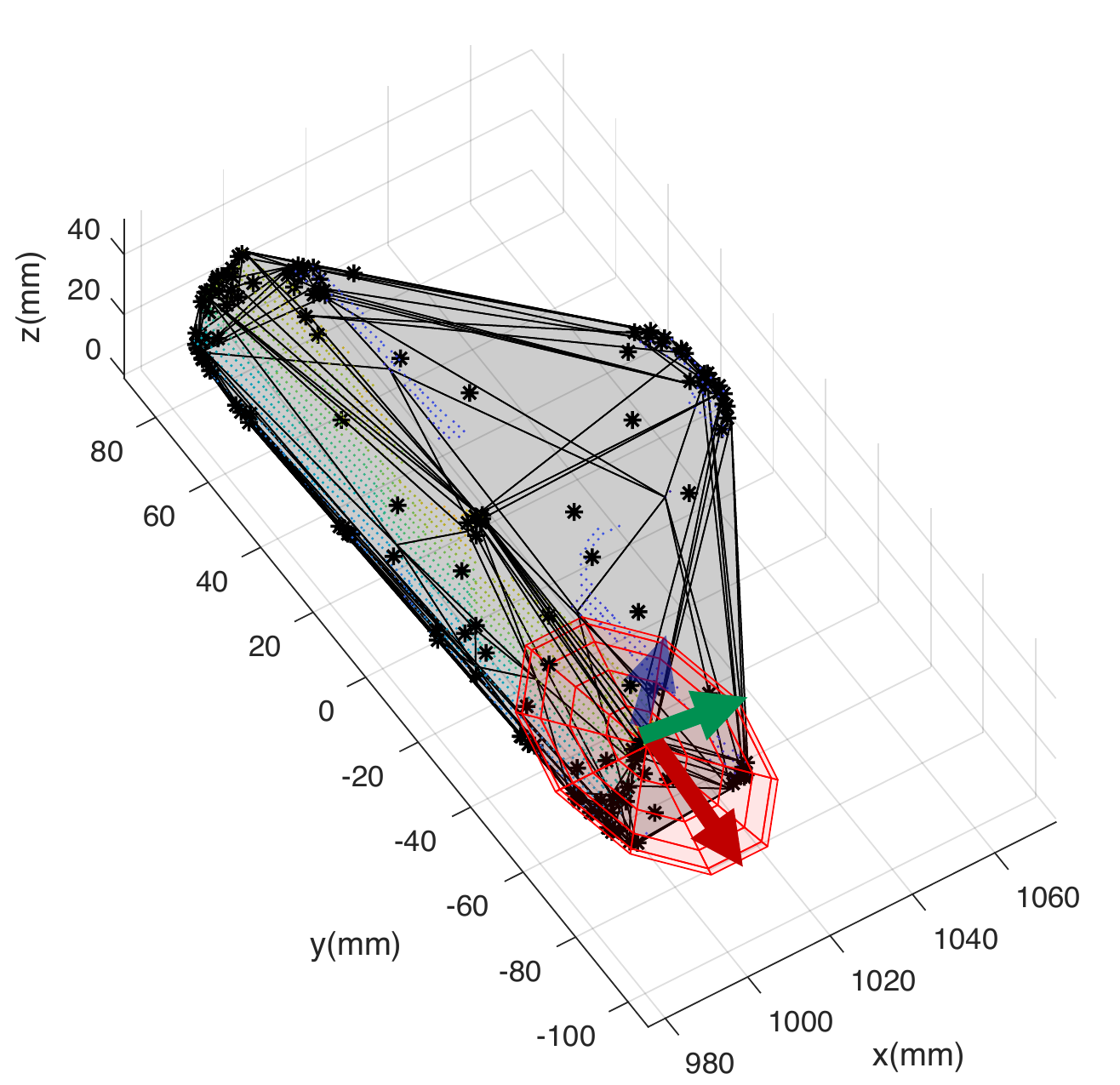}
            \end{minipage}} 
            \subfigure[to contain]{ \label{fig:glass}
                \begin{minipage}{0.18\linewidth}
                    \centering
                    \includegraphics[width=2cm]{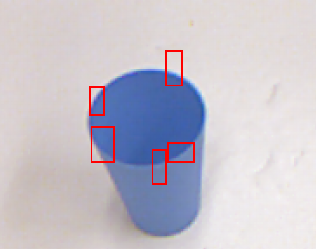}
                    
                    \includegraphics[width=2.2cm]{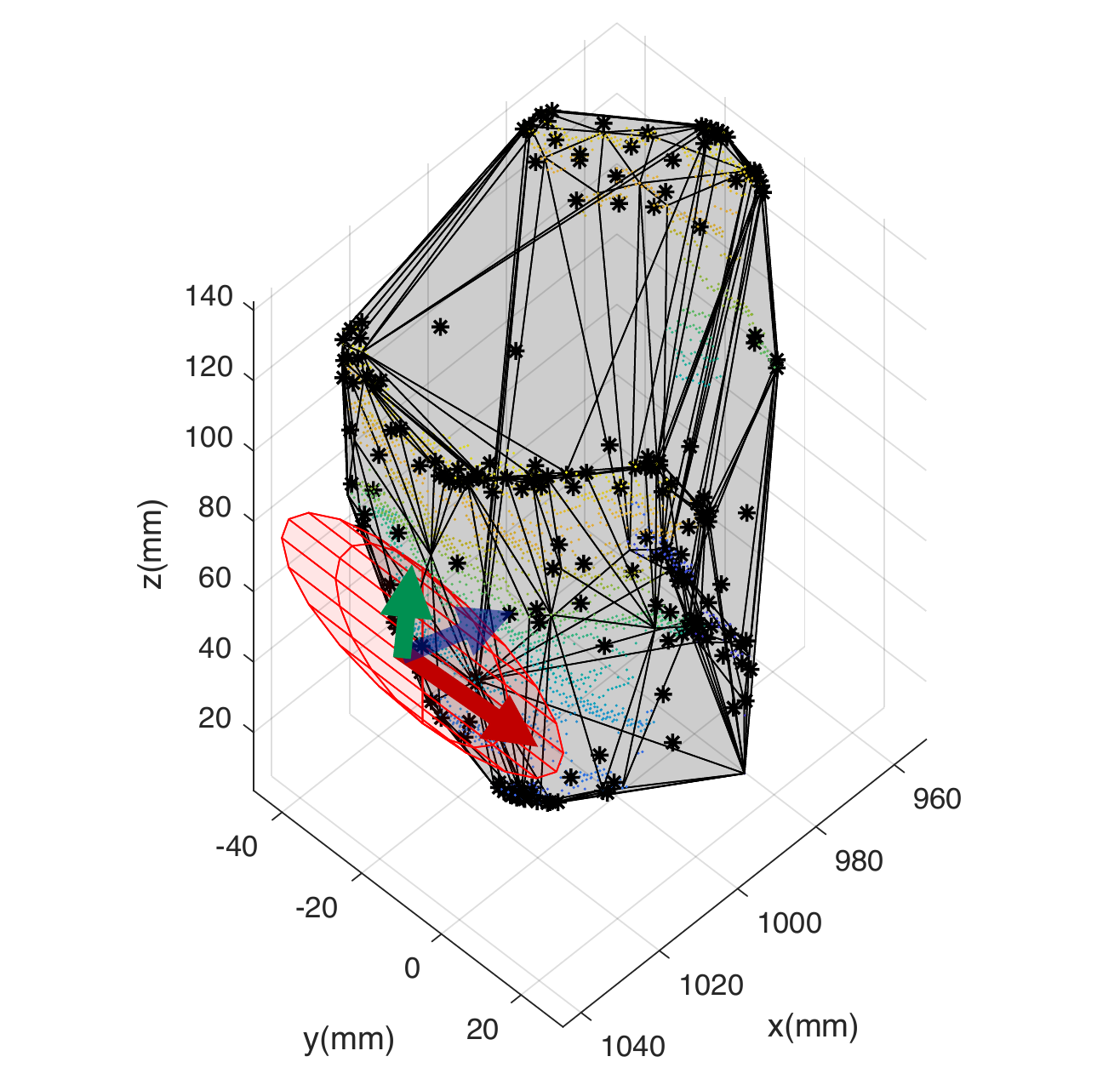}
                \end{minipage}} 
            \subfigure[to hand over]{
                \begin{minipage}{0.18\linewidth}
                    \centering
                    \includegraphics[width=2cm]{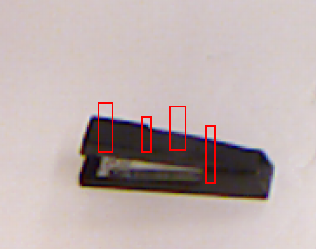}
                    
                    \includegraphics[width=2.9cm]{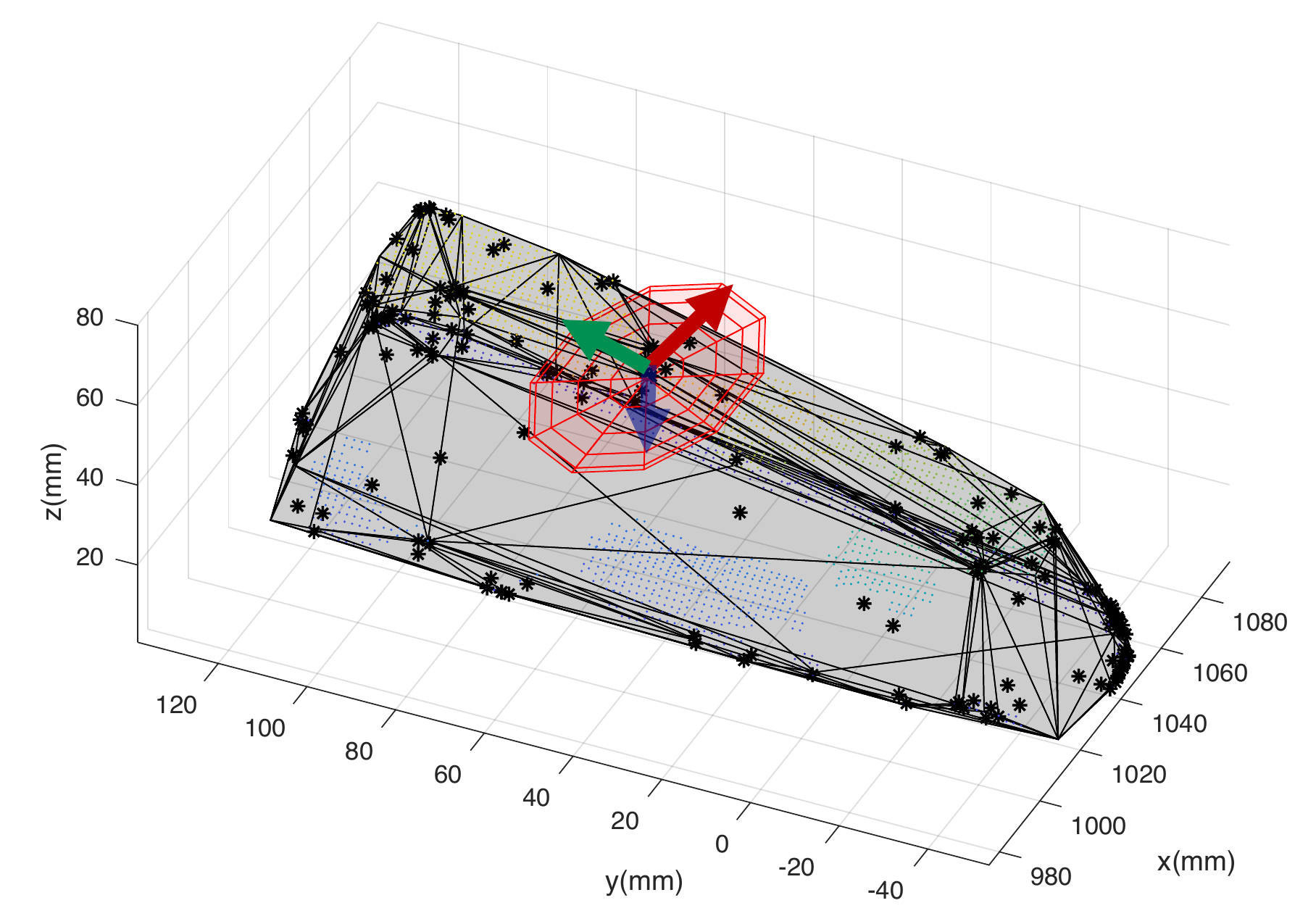}
                \end{minipage}} 
            \subfigure[to contain]{ \label{fig:mug}
                \begin{minipage}{0.18\linewidth}
                    \centering
                    \includegraphics[width=2cm]{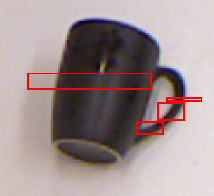}
                    
                    \includegraphics[width=2.2cm]{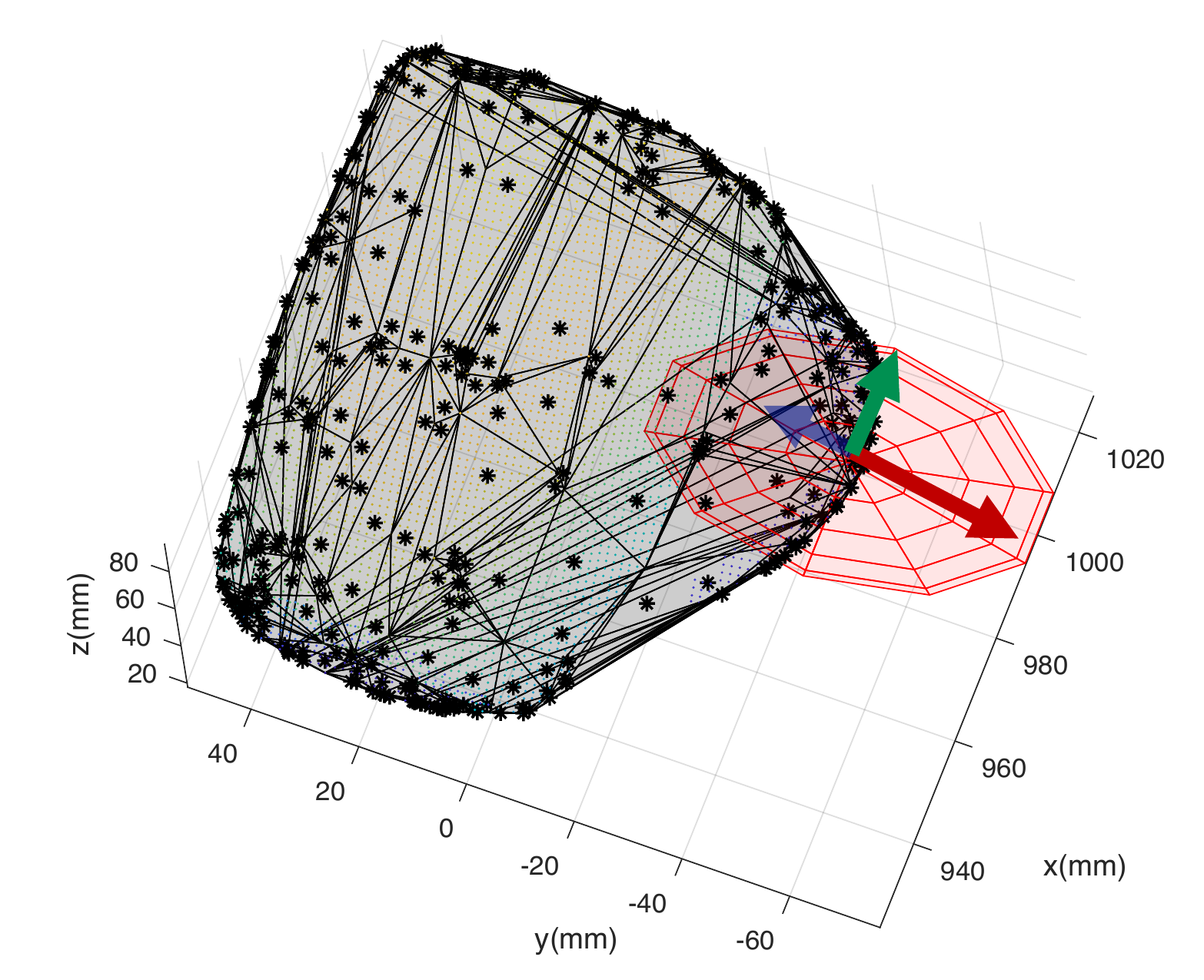}
                \end{minipage}} 
            \subfigure[to hand over]{\label{fig:candle}
                \begin{minipage}{0.18\linewidth}
                    \centering
                    \includegraphics[width=2cm]{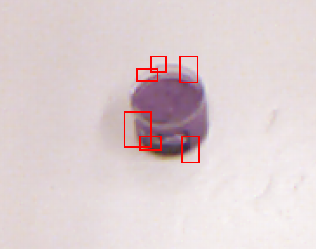}
                    
                    \includegraphics[width=2.6cm]{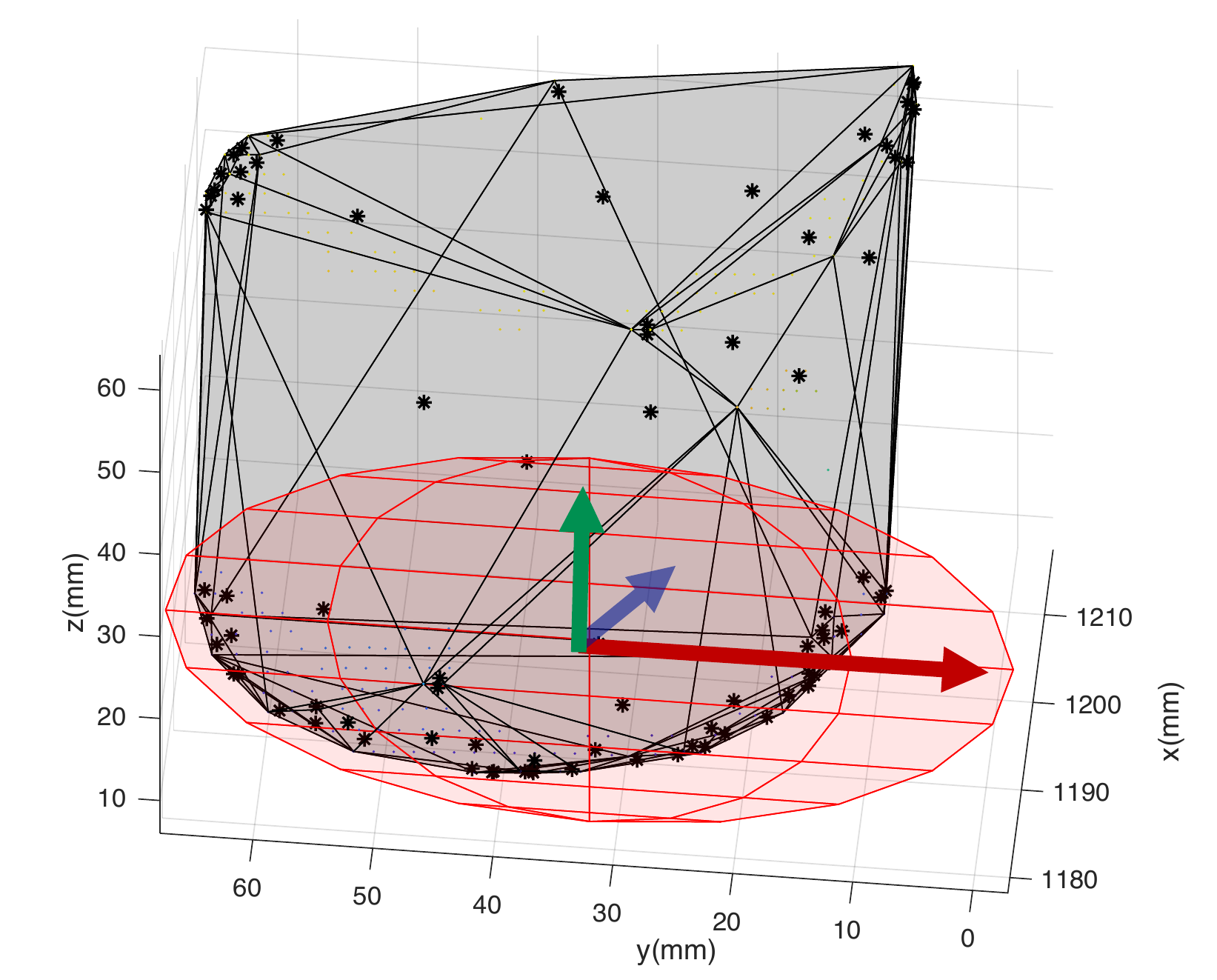}
                \end{minipage}} 
            \caption{Zero-shot affordance prediction on semantically similar objects. The original images contain the labels (rectangles) for the preferred grasping regions from \cite{Lenz2015,sung2017deep}.\label{fig:zero}}
            \vspace{-0.5cm}
    \end{figure*}
%----------------------

Considering the changing nature of indoor scenes, it is useful to measure the method's affordance prediction on new objects.
In this work, the object affordance is limited to its grasping action and is seen as the combination of the action-effect pair that results from the observations of the object and its environment. Zero-shot affordance, in this case, refers to the affordance prediction of a familiar but previously unseen object. For this part of the experiments, a set of semantically similar objects has been chosen from a third dataset, Cornell \cite{sung2017deep}. This dataset is used to learn how to grasp objects in other works such as \cite{Lenz2015,sung2017deep}. 
These works exploit the fact that the Cornell dataset contains the \ac{3-D} point cloud of the objects and their corresponding labelled grasping regions in the form of rectangles. From the Cornell dataset, 22 semantically similar objects to the ones used for the training of the \ac{KB} are chosen, obtaining an average accuracy of 81.3\% on the object affordance reasoning. In order to deduce the affordance of an unknown object, the same hierarchical procedure previously explained is followed. The set of weights $\boldsymbol \Psi_{A}$ has ranked a connection of attributes that results in an affordance, depending on the perceived semantics. Furthermore, this hierarchical connection has been learned in a predictive model to result in the grasping areas of the object. Figure~\ref{fig:zero} shows a sample of the familiar objects tested using the \ac{KB} with their affordance group and deduced grasping area (shown with the red ellipse). Out of this subset, the most critical case is shown by the ones which affordance is to contain edibles, the cup and the mug in Figures~\ref{fig:glass} and \ref{fig:mug}, for which the grasping areas are correctly calculated.

%%%%%%%%%%%%%%%%%%%%%%%%%%%%%
%%%%%%%%%%%%%%%%%%%%%%%%%%%%%
%%%%%%%%%%%%%%%%%%%%%%%%%%%%%
\section{Discussion}

The proposed methodology is not only able to (i)~reason on the object affordance of known and semantically similar objects, but also~(ii)~to extract a suitable grasp-action region of the target depending on the interpreted affordance. Given these features, this section discusses the performance of the \ac{KB} on discerning the affordance of semantically similar objects, followed by a comparison of the obtained grasp-action regions with other methods' ground truth data.

%-------------------------
%-------------------------
%-------------------------
\vspace{-0.3cm}
\subsection{Similar Shape, Different Affordance}
\vspace{-0.3cm}
One of the most significant arguments for building this framework is to help a robot generalise on object affordances. That is to say, just as humans succeed at generalising an action towards objects of the same category with significantly different shapes, e.g. glasses: wine, tumbler, martini, and differentiate how to manipulate objects with similar shapes but for different purposes, e.g. candle vs water bottle. Given the objects in the library, it is of interest to evaluate the different affordance and grasping regions obtained for objects with similar shape but different affordance thus different preferred grasping regions. Figure~\ref{fig:glass} and Figure~\ref{fig:candle} are examples of two different everyday objects (a cup and a candle respectively) with considerably different affordance, where the located grasping regions differ according to the deduced affordance of the object.

%-------------------------
%-------------------------
%------------------------
\vspace{-0.3cm}
\subsection{Quality on the Calculated Grasping Area}
\vspace{-0.3cm}
Different works have been done in the field of affordance detection and grasping. However, they commonly learn a labelled set of data in order to be able to identify the grasping regions. Contrary to these techniques, the method presented in this paper deduces the grasping region without any \textit{a-priori} information about the grasping points.
Given that the presented method does not train on grasp labels, in order to evaluate its output, it is compared to the ground truth labels of the Cornell dataset. There are works that use deep learning techniques to learn the grasping points of the objects mapped in the Cornell dataset images \cite{Lenz2015,sung2017deep,Saxena2008RoboticGO}. It is worth mentioning that these works do not account for affordance learning but for object classification. They simulate the end-effector with a rectangle, allowing it to account for its orientation, and use point and rectangle metrics to measure the \ac{MSE} between their ground truth and the obtained grasps. 
Their proposed point metric computes the centre point of the predicted rectangle and considers the grasp as a success if it is within some distance from at least one of the ground truth rectangles. Contrary to this work, their labelled grasping areas are based on their end-effector control, and kinematic constraints and not on object affordance. Thus, a direct quantitative comparison is not viable.
However, it is possible to use a modified version of their proposed point metric. The results of this work can be qualitatively evaluated by visually inspecting the resulting area. Moreover, quantified by the percentage of grasping regions that coincide between both sets of data, i.e., the labelled rectangles of the Cornell dataset and the ellipses of this proposal.
In order to obtain such percentage, the Euclidean distance from the centre point of the labelled rectangles, observation $a$, to the centre point of the superellipsoid, observation $b$, is measured and expected to be below a set threshold. From the Cornell dataset, a subset of 65 random images was taken, including images from different perspectives of the same object. These images were categorised into an affordance group, illustrating their provided grasping label as a red rectangle on the \ac{2-D} image, as seen in Figure~\ref{fig:zero}. By measuring the Euclidean distance, 88\% of the calculated grasps using the \ac{KB} proposed in this work fall inside the labelled grasping regions. The other 12\% falls either close to a valid region, or entirely in a new area given that it has followed the constraints of the grasps depending on the object affordance, as it is the case of the cup in Figure~\ref{fig:glass}. 

\section{Conclusions and Future Work}\label{sec:contribution}

% The literature presents many solutions for object affordance and grasping behaviours. Nevertheless, considering the object's grasping areas as the result of its affordance is still an open challenge. An usual approach in literature is to take advantage of labelled data to deal with the extraction of grasping regions. However, this technique limits the number of objects, grasping areas and categories a system can learn thus not being generalisable. 

Contrary to the available methods, the framework presented in this paper is able to (i)~reason on the affordance grasp-action of known and familiar objects without previously acknowledging the grasping areas, thus (ii)~offering a reasoning process for object interaction with autonomy capabilities. The results of the evaluation performed on the framework support the hypothesis presented at the beginning of this work: that the grasp-action affordance does not depend solely on the object semantic features but on their combination with the features that describe the environment. 
The results show that without any \textit{a-priori} awareness on the grasping regions, the designed \ac{KB} can reason on the object's affordance grasping points.
% Furthermore, by building a \ac{KB} the system does not only learn the final predictive affordance model, but it can also access high-level information that allows it to distinguish different visual semantic attributes of the objects and their related environment. 
% Moreover, a comparative evaluation was carried out considering the ground truth used in methods that apply learning techniques to extract the grasping regions. These techniques, contrary to the one presented in this paper, depending on the provided labelled data in order to obtain the grasping areas. By comparing the previously unseen labelled grasping areas, the proposed method in this paper obtains an 81.3\% accuracy at reasoning the affordance of semantically similar objects and 88\% of matching areas with the labels. The outcome of this comparison positively asserts the application of the \ac{KB} for object interaction in indoor environments.
The presented framework has room for improvement. The performance of the \ac{KB} can be increased by adding more attributes to the base, as well as modifying the predictive model to classify more than one affordance at the time (for example, an object's affordance can be \textit{to hand over} as well as \textit{to clean}).
Furthermore, the dynamics and system control schemes of the robot and the environment are considered out of the scope of the presented work. Nonetheless, \cite{pairet2018learning,pairet2019learning} offers a learning-based framework that comprises relative and absolute robotic skills for dual-arm manipulation suitable for dynamic environments, that together with a dense context representation of the scenario semantics offers a complete solution for an interactive object platform.
\section{ACKNOWLEDGEMENTS}
\vspace{-0.3cm}
Thanks to the support of the EPSRC IAA 455791 along with ORCA Hub~EPSRC (EP/R026173/1, 2017-2021) and consortium partners.

\bibliographystyle{splncs04}
\bibliography{main.bbl}
\end{document}